\pdfoutput=1

\documentclass[11pt]{article}

\usepackage[]{acl}

\usepackage{times}
\usepackage{latexsym}

\usepackage[T1]{fontenc}

\usepackage[utf8]{inputenc}

\usepackage{microtype}

\usepackage{times}

\usepackage{soul}
\usepackage{url}
\usepackage[utf8]{inputenc}
\usepackage{graphicx}
\usepackage{amsmath}
\usepackage{booktabs}
\usepackage{algorithm}
\usepackage{algorithmic}
\usepackage{enumerate}
\usepackage{amsmath}
\usepackage{bm}
\DeclareMathOperator*{\argmax}{argmax}
\usepackage{makecell}
\usepackage{multirow}
\usepackage{booktabs}
\usepackage{arydshln}

\title{ArT: All-round Thinker for Unsupervised Commonsense \\ Question Answering}

\author{Jiawei Wang\textsuperscript{1,2}, Hai Zhao\textsuperscript{1,2\thanks{\ \ Corresponding author. This paper was partially supported by Key Projects of National Natural Science Foundation of China (U1836222 and 61733011).}}\\
\textsuperscript{1} Department of Computer Science and Engineering, Shanghai Jiao Tong University\\
\textsuperscript{2} Key Laboratory of Shanghai Education Commission for Intelligent Interaction\\
and Cognitive Engineering, Shanghai Jiao Tong University\\
\texttt{wjw\_sjt@sjtu.edu.cn, zhaohai@cs.sjtu.edu.cn}\\}

\begin{document}
\maketitle
\begin{abstract}
Without labeled question-answer pairs for necessary training, unsupervised commonsense question-answering (QA) appears to be extremely challenging due to its indispensable unique prerequisite on commonsense source like knowledge bases (KBs), which are usually highly resource consuming in construction. Recently pre-trained language models (PLMs) show effectiveness as an alternative for commonsense clues when they play a role of knowledge generator. However, existing work either relies on large-scale in-domain or out-of-domain labeled data, or fails to generate knowledge of high quality in a general way. Motivated by human thinking experience, we propose an approach of \textbf{A}ll-\textbf{r}ound \textbf{T}hinker (ArT) by fully taking association during knowledge generating. In detail, our model first focuses on key parts in the given context, and then generates highly related knowledge on such a basis in an association way like human thinking. Besides, for causal reasoning, a reverse thinking mechanism is especially added to further enhance bidirectional inferring between cause and effect. ArT is totally unsupervised and KBs-free. We evaluate it on three commonsense QA benchmarks: COPA, SocialIQA and SCT. On all scales of PLM backbones, ArT shows its brilliant performance and outperforms previous advanced unsupervised models. Our code is available at \url{https://github.com/WangJW424/commonsenseQA-ArT}.
\end{abstract}

\section{Introduction}

\begin{figure}[htb]
	\centering
	\includegraphics[width=0.48\textwidth]{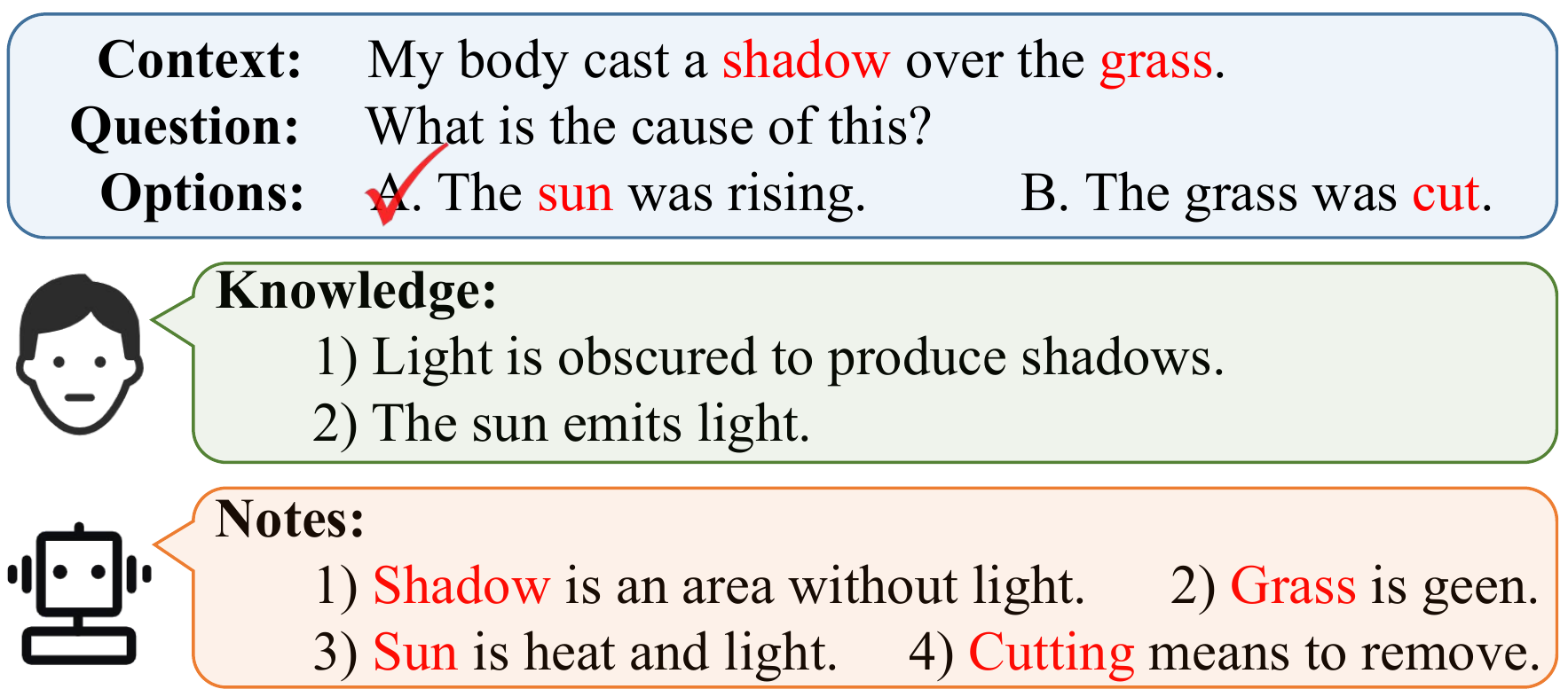}
	\caption{An commonsense QA exmaple in COPA. Text with blue background is the raw example and the gold answer is ticked. Text with green background is the needed commonsense knowledge. Text with orange background is notes generated by our ArT. Keyphrases are marked in red.}
	\label{fig:commonsenseqa_example}
    \end{figure}
    Commonsense question-answering (QA) has been a more challenging natural language understanding (NLU) task than conventional QA tasks, for it requires extra commonsense knowledge, which cannot be directly acquired from the given context, to make an appropriate answer \cite{niu2021semantic}. Figure \ref{fig:commonsenseqa_example} gives a typical example. To select the correct cause of the fact \textit{``My body cast a shadow over the grass''}, models should know that \textit{``Light is obscured to produce shadows''} and \textit{``The sun emits light''}. As human beings, we can immediately make this association and judgment because above knowledge is deeply branded in our mind due to daily seeing and hearing, and maybe long-term education, which is so-called commonsense. However, it could be much difficult for models to be equipped with such ability. Besides, commonsense QA usually takes an unsupervised setting which makes this task even more difficult. This setting means there is no labeled training data available as commonsense is too broad to build a sufficient labeled training dataset \cite{shwartz2020unsupervised}.
    
    To deal with it, prior work focused on building large-scale knowledge bases (KBs), also known as knowledge graphs (KGs), which usually contain millions of nodes and edges to record the relation between entities as relation triple: $<$$e_1, r, e_2$$>$ \cite{speer2017conceptnet,sap2019atomic}. QA models then can be injected with commonsense through retrieving over KBs \cite{miller2016key}. Though impressive improvements are gained, such method is resource consuming in building or finding a good and suitable KB (e.g. Building ConceptNet needs 30 GB of RAM\footnote{Declared by \citet{speer2017conceptnet} at \url{https://github.com/commonsense/conceptnet5}}). Recently, pre-trained language models (PLMs) have been widely used and proved to be effective in commonsense QA \cite{niu2021semantic, xia2022prompting}. Thanks to the self-supervised pre-training strategy on large-scale unlabeled text (e.g. WebText \cite{radford2019language} and Wikipedia\footnote{https://www.english-corpora.org/wiki}), PLMs are competent for many tasks even under a zero-shot setting. And fine-tuning PLMs on task-specific data in a supervised way can further produce even stronger results \cite{schick2021exploiting, gao2021making, schick2021s}. Since high-quality labeled datasets are rare, researching on unsupervised commonsense QA is still of great significance. With the help of PLMs, existing studies have explored some good solutions \cite{niu2021semantic,bosselut2021dynamic,shwartz2020unsupervised}, however they either rely on large-scale in-domain or out-of-domain labeled data, or need to be specifically designed for different tasks. In this work, we focus on designing a simple and general method to solve commonsense QA tasks in a strictly unsupervised way.
    
    Based on two empirical observations from human thinking,
    \begin{enumerate}[(1)]
        \item Given a question with context, people firstly tend to focus on several key parts (as marked in red in Figure \ref{fig:commonsenseqa_example}) and then make corresponding associations to choose the right answer;
        \item For causal reasoning, people tends to carry out a bidirectional inferring to assist answer selection or verify the answer correctness. 
    \end{enumerate}
    we propose \textbf{A}ll-\textbf{r}ound \textbf{T}hinker (ArT) for unsupervised commonsense QA, which includes two principal methods: notes taking (NT) and reverse thinking (RT). Specifically, 
    \begin{enumerate}[(1)]
    \item NT extracts some keyphrases out of the context and then generates corresponding notes, which will be added as extra knowledge in later evaluation. Based on an unsupervised keyphrase extractor, we designed our knowledge generation rule to be simple and general.
    \item RT converts the causal inferring question to two different forms: (\textit{cause} $\to$ \textit{effect}) and (\textit{effect} $\to$ \textit{cause}), and then integrates the decisions made from the two reverse directions. 
    \end{enumerate}
    
    Our proposed model is strictly unsupervised and KBs-free for all it needs is PLMs. We test ArT and validate its effectiveness on three commonsense QA benchmark datasets: COPA \cite{roemmele2011choice}, SocialIQA \cite{sap2019social} and SCT \cite{mostafazadeh2016corpus}. Our contribution is summarized as follows:
    \begin{itemize}
        \item ArT can generate highly related knowledge through the imitation of human behaviour and thought, which is qualified with inherent interpretablility.
        \item Compared with existing work, ArT is simple and general, which totally gets rid of the needs of any labeled data and the specific design on any specific task.
        \item We conduct experiments on 3 commonsense QA benchmarks with 4 different scales of PLMs, so as to reach solid and reproducible results. The results show that ArT outperforms other advanced unsupervised models.
    \end{itemize}

\begin{figure*}[ht!]
	\centering
	\includegraphics[width=0.92\textwidth]{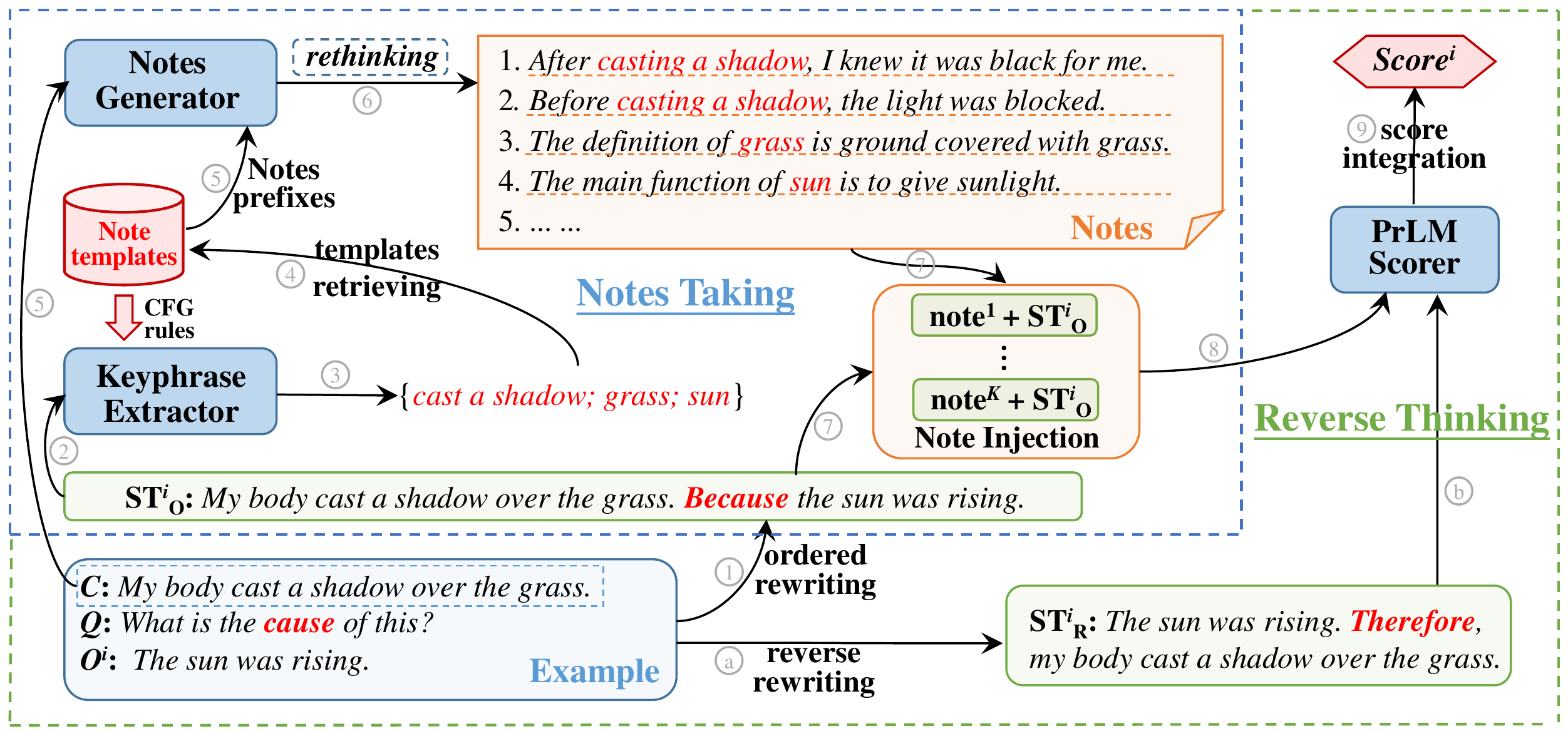}
	\caption{Overview of the proposed method \textbf{A}ll-\textbf{r}ound \textbf{T}hinker (ArT), which contains two principal method: Notes Taking (within blue dashline box) and Reverse Thinking (within green dashline box). The arrow denotes the flow of data stream, with number/letter within circle marking its order in time.}
	\label{fig:overview}
\end{figure*}

\section{Related Work}
\subsection{Building and Usage of Knowledge Bases}
    In order to equip QA models with commonsense reasoning ability, previous researches were devoted to building and retrieving large-scale knowledge bases (KBs). ConceptNet \cite{speer2017conceptnet} is one of the most famous traditional KBs, which contains over 21 million edges (for relations) and over 8 million notes (for entities). While ATOMIC \cite{sap2019atomic} focuses more on \textit{if-then} relations between events. 
    
    Previous work applies a relatively standard routine to solve commonsense QA. Given an existing KB, QA models can retrieve relation triples $<$$e_1, r, e_2$$>$ over it, which can be injected into models as word embedding directly \cite{wang2014knowledge,paul2019ranking} or first converted to a complete sentence according to preset templates (e.g. $<$bird, \textit{CapableOf}, fly$>$ $\to$ \textit{Bird can fly.}) and then integrated with the raw input text \cite{ma2019towards,mihaylov2018knowledgeable,bauer2018commonsense}. This type of methods may give remarkable performance for commonsense QA \cite{weissenborn2017dynamic}. However, building and retrieving of KBs are both resource consuming \cite{bosselut2021dynamic}.
    
    \citet{bosselut2019comet} claimed that commonsense knowledge does not cleanly fit into a schema of comparing two entities with a known relation so that they proposed COMET, which can generate rich and diverse commonsense descriptions in natural language. However, COMET has to extract knowledge triples from existing KBs as seed for PLM fine-tuning, while our method is totally KBs-free.

\subsection{PLMs in Unsupervised Commonsense QA}
    Due to excellent performance and versatility, PLMs now dominate the backbone design of many NLP tasks, especially in QA \cite{wang2021sentence-hood, zhang2021retrospective}. Benefiting from the long-term pre-training on large-scale unlabeled text, PLMs are equipped with implicit or explicit commonsense \cite{davison2019commonsense}. As a result, there comes a tendency to rely on PLMs as the sole source of world knowledge to solve commonsense QA in a zero-shot setting \cite{shwartz2020unsupervised,bosselut2021dynamic}. 
    
    SEQA proposed by \citet{niu2021semantic} exploits generative PLMs to generate hundreds of pseudo-answers, and then applies Sentence-BERT \cite{reimers2019sentence} to calculate semantic similarity between candidates and these generated pseudo-answers. A voting mechanism is designed to make final selection. Though SEQA shows impressive performance, it relies much on PLMs fine-tuned on large-scale labeled NLI (Natural Language Inference) datasets. Without such fine-tuning on such labeled datasets, SEQA's effectiveness will sharply decline. Instead, our method relies nothing but vanilla PLMs.
    
    \citet{shwartz2020unsupervised} proposed Self-talk, which creatively applies PLMs as commonsense generator. It uses preset question prefixes to firstly prompt the PLM to generate information seeking questions (ISQs). ISQs will be put back to secondly prompt the PLM to generate their answers as clarifications, which will work as knowledge in later evaluation. Though this method is strictly unsupervised, the question prefixes are not general and have to be carefully designed according to different tasks. Besides, Self-talk fails to generate knowledge of high quality since the generation of ISQs and clarifications are both erratic. Inspired by Self-talk and towards its shortage, we designed a general method to generate knowledge of high relevance.

\subsection{Unsupervised Keyphrase Extraction}
    Keyphrase extraction (KE) is the task of selecting several words or phrases that can summarize the main topic of the document \cite{hasan2014automatic}. Unsupervised KE methods use different features of the document, such as word frequency, position feature, relationship between words, etc \cite{mihalcea2004textrank,bougouin2013topicrank}.
    
    SIFRank \cite{sun2020sifrank} is one of the most advanced unsupervised keyphrase extractors. It leverages different features of words in document and evaluates the weight of candidate phrases according to word embedding provided by PLMs. SIFRank is originally designed to extract key noun phrases from given documents, and we made slight modifications to adapt it to our model.

\begin{table*}[ht!]
    \fontsize{10pt}{\baselineskip}
    \selectfont
	\centering
 	\resizebox{\linewidth}{!}{
	\begin{tabular}{c|ll|c}  
	\Xhline{0.8pt}
	\textbf{Key} & \multicolumn{2}{c|}{\textbf{Value}} & \textbf{Replacing rule} \\
	\hline
    ``NP'' & \{ ``\textit{The definition of \textup{[NP]} is}'', ``\textit{The main function of \textup{[NP]} is}'', ``\textit{\textup{[NP]} is a/an}'' \} & & directly replace \\
    ``VP'' & \{ ``\textit{\textup{[VP]} means}'', ``\textit{After \textup{[VP]}, }'', ``\textit{Before \textup{[VP]}, }'' \} & & convert to gerund first\\
    ``PNP'' & \{ ``\textit{\textup{[PNP]} is a/an}'', ``\textit{\textup{[PNP]} felt}'', ``\textit{After this, \textup{[PNP]}}'', ``\textit{\textup{[PNP]} did this because}'' \} & & directly replace\\
    \bottomrule
	\end{tabular}
	}
	\caption{Note templates lookup table.}
	\label{tab:dict}
    \end{table*}
    
\section{All-round Thinker}
    Inspired by the behaviour and thought of human beings during solving questions requiring commonsense, we propose \textbf{A}ll-\textbf{r}ound \textbf{T}hinker (ArT). Figure \ref{fig:overview} gives its overview. We will firstly introduce the task definition and the basic solution, and then describe the two principal methods of ArT in detail: Notes Taking (NT) and Reverse Thinking (RT).

\subsection{Task Definition and Basic Solution}
\label{basic}
    This work focuses on unsupervised commonsense QA task in multi-choice style, which consists of a $context$ ($C$) with related $question$ ($Q$), and asks models to select a single answer from a given $option$ set: $O=\{O^i\}_{i=1}^{|O|}$. Though there are some variants of the task form, such as Piqa \cite{bisk2020piqa} and WinoGrande \cite{sakaguchi2020winogrande}, they can be conveniently transformed into this normalized formulation: ($C$, $Q$, $O$) \cite{shwartz2020unsupervised}.

    Following previous work, we adopt a basic solution which uses PLMs as scorer. Based on the pre-training strategy: predicting the probability distribution of token $n$ according to previous $(n-1)$-gram, e.g. OpenAI GPT \cite{radford2019language}, or bidirectional context, e.g. BERT \cite{devlin2019bert}, PLMs are qualified for the role of $option$ scorer even under a zero-shot setting \cite{bosselut2021dynamic}. Sentence likelihood is the commonly-used scoring function for $O^i$:
    \begin{equation}
        \begin{split}
            \bm{S}(O^i|C, Q)&=P_{LM}(O^i|C, Q) \\
                  &=\frac{1}{|O^i|}\sum\limits_{t=1}^{|O^i|} \log P_{LM}(O_t^i|C, Q, O_{<t}^i)
        \end{split}
    \label{eq:1}
    \end{equation}
    where $P_{LM}$ refers to the probability function abstracted from PLMs. The final predicted answer ($\hat A$) is selected as:
    \begin{equation}          
        \hat A = \argmax\limits_{{O^i}}\bm{S}(O^i|C, Q)
    \label{eq:2}
    \end{equation}

\subsection{Notes Taking}
\label{sec:nt}
    To make full use of PLMs' potential of commonsense reasoning and overcome the shortage of Self-talk \cite{shwartz2020unsupervised}, we propose notes taking (NT) to generate commonsense descriptions in natural language (defined as ``notes'' in our work) in a simple and general way. NT is composed of two phases: keyphrase extraction and notes generation.

\subsubsection{Keyphrase Extraction}
    \label{sec:ke}
    Taking $O^i$ and its corresponding $C$ and $Q$ as input (as shown in left bottom of Figure \ref{fig:overview}), ArT firstly rewrites the original interrogative $Q$ into a declarative form (We followed the question rewriting method proposed by \citet{shwartz2020unsupervised}) and then concatenate $<$$C$, $Q$, $O^i$$>$ as a statement ($ST^i_O$). Then, we use an unsupervised keyphrase extractor to extract keyphrases from $ST^i_O$. To be specific, we extract three types of phrases:  noun phrase (NP), verb phrase (VP) and person name phrase (PNP). 
    
    To implement this, for each type of phrase we designed a simple CFG (context-free grammar \cite{chomsky1959algebraic}) rule to extract it out of the whole sentence, as following:
    \begin{itemize}
        \item $NP \to (nn|adj)*+ nn$
        \item $VP \to vb+(pr)\{0,1\}+NP$
        \item $PNP \to (pn)\{1,2\}$
    \end{itemize}
    where \textit{VP}, \textit{NP} and \textit{PNP} are non-terminators; $vb$ (verb), $nn$ (noun), $adj$ (adjective), $pn$ (person name) and $pr$ (preposition) are terminators; `$+$' means concatenation; $\{a, b\}$ means repetition times range from $a$ to $b$. `$*$' is equivalent to $\{0, \infty\}$.
    
    Then, we add these CFG rules to the RegexpPaser tool of NLTK \cite{bird2006nltk}. We will extract top 5 most important keyphrases without any label or fine-tuned model but word embeddings obtained from a PLM, i.e. ELMo \cite{peters2018deep}.
    
    \subsubsection{Notes Generation}
    Once keyphrases are obtained, we can retrieve a preset note templates set for getting notes prefixes. Considering that: (1) For an object (NP), people will think “what is it” and “what is it for”; (2) For a behavior (VP), people will think “what does it mean” and “what is the cause/effect”; (3) For a person (PNP), people will think “who is he/she” and “what is his/her feeling/motivation/reaction”, our proposed general-purpose note templates set is presented in Table \ref{tab:dict}. 
    
    We use the types of keyphrases as $keys$ and templates lists as $values$ to build the note templates set as a lookup table. Given such a lookup table, we can immediately retrieve the note templates for our extracted keyphrases and simply replace the tag ([NP], [VP] and [PNP]) with these keyphrases to form the note prefixes\footnote{In order to keep a correct grammar, for VP we will firstly convert its verb into gerund form before replacing.}. Though this lookup table seems to be simple, it shows effectiveness and generality on different benchmarks.
    
    Next, for each note prefix, we will concatenate it to $C$ and input them into a generative PLM to generate the complete note. Specifically, nucleus sampling \cite{holtzman2020curious} with $p = 0.8$ is applied as the decoding strategy rather than greedy/beam search to increase the \textbf{diversity} of generated text. Meanwhile, to ensure the \textbf{quality} and scale the number of generated notes, we sort all the notes according to their perplexity estimated by the PLM. We denote this process as \textbf{\textit{Rethinking}}. Finally, we will retain top $K$ notes to construct the notes set: $\{note^k\}_{k=1}^K$, as shown in middle top of Figure \ref{fig:overview}. Each $note^k$ will be inserted into $ST^i_O$ as extra knowledge to assist later \textit{option} scoring. The score of $O^i$ w.r.t $note^k$ is calculated as:
    \begin{equation}
        \begin{split}
            &score^{i,k}_O=P_{LM}(O^i|note^k+ST^i_O-O^i) \\
                         &=\frac{1}{|O^i|}\sum\limits_{t=1}^{|O^i|} \log P_{LM}(O_t^i|note^k+ST^i_O-O^i+O_{<t}^i)
        \end{split}
    \label{eq:3}
    \end{equation}
    in which `+' means concatenation and `-' means removing. Eventually, a voting mechanism is applied to integrate the scores w.r.t all notes as:
    \begin{equation}
        \begin{split}
            Score^i_O&=\frac{1}{K}\sum\limits_{k=1}^{K}score^{i,k}_O
        \end{split}
    \label{eq:5}
    \end{equation}
    
    It is worth noting that there is no need in training from any labeled data throughout NT. And there is no need to modify the note templates according to different tasks since our note templates is designed for general purpose.
    
    \subsection{Reverse Thinking}
    \label{sec:rt}
    For causal reasoning questions, we additionally introduce reverse thinking (RT) which conducts a bidirectional inferring between \textit{cause} and \textit{effect}.
    To implement this, besides the ordered rewriting ($ST^i_O$) as mentioned in Section \ref{sec:ke}, we also apply reverse rewriting that concatenates them in the order of $<$$O^i$, $Q_R$, $C$$>$ (denoted as $ST^i_R$), as shown in the right bottom of Figure \ref{fig:overview}. Note that $Q_R$ is the opposite question of $Q$. To be specific, after question rewriting, ``\textit{Because}'' and ``\textit{Therefore}'' are two opposite questions in causal reasoning tasks.
    
    To conduct bidirectional inferring, except for $Score^i_O$ as introduced in Section \ref{sec:nt} for the ordered inferring: $C+Q\to O^i$, we set another scoring function $Score^i_R$ for reverse inferring: $O^i+Q_R\to C$, as:
    \begin{equation}
        \begin{split}
            Score^i_R&=P_{LM}(C|ST^i_R-C) \\
                         &=\frac{1}{|C|}\sum\limits_{t=1}^{|C|} \log P_{LM}(C_t|ST^i_R-C+C_{<t})
        \end{split}
    \label{eq:4}
    \end{equation}
    
    To take advantage of bidirectional inferring, we design a mixed scoring function by simply compute the average value of the above two:
    \begin{equation}
            Score^i_X=\frac{1}{2}(Score^i_O + Score^i_R)
    \label{eq:6}
    \end{equation}
    Finally, formula (\ref{eq:2}) is applied to select the answer by replacing $\bm{S}$ with $Score^i_O$ (default) or $Score^i_X$ (for causal reasoning). From the perspective of model enhancing, averaging $Score^i_O$ and $Score^i_R$ can be regarded as the assembly of two models (``cause'' model and ``effect'' model), which is a common method to enhance model performance and robustness.

\section{Experiment}
\label{sec:experiment}
\subsection{Datasets}
    ArT is evaluated on three different commonsense QA benchmarks: COPA\cite{roemmele2011choice}, SocialIQA\cite{sap2019social} and SCT\cite{mostafazadeh2016corpus}. Here are the detailed information:
    \begin{itemize}
        \item \textbf{COPA}\footnote{\url{https://people.ict.usc.edu/~gordon/copa.html}} (\textbf{C}hoice \textbf{o}f \textbf{P}lausible \textbf{A}lternatives): evaluates the ability of causal reasoning about a certain event, which is described as a single sentence. Each question is accompanied with two candidate options.
        \item \textbf{SocialIQA}\footnote{\url{https://leaderboard.allenai.org/socialiqa}} (\textbf{Social} \textbf{I}nteraction \textbf{Q}uestion \textbf{A}nswering): evaluates the reasoning ability on social interactions. It has various questions, including the subject's motivation, reaction, personality, etc. Each question is accompanied with three candidate options. 
        \item \textbf{SCT} \footnote{\url{https://www.cs.rochester.edu/nlp/rocstories/}} (\textbf{S}tory \textbf{C}loze \textbf{T}est): requires models to select the right ending of the given short story from two alternatives. Each story is composed of four sentences. 
    \end{itemize}
    
    Since two test sets of the three datasets are hidden, we report all results on dev sets. Note that the labels are kept invisible and only used for final accuracy evaluating. 

\begin{table*}[ht!]
    \fontsize{10pt}{\baselineskip}
    \selectfont
	\centering
	\begin{tabular}{llllll|c}  
	\toprule
	& & \multicolumn{4}{c|}{\textbf{Our (re-)running}} & \textbf{Published} \\
	\textbf{Dataset} & \textbf{Models} &  DistilGPT-2 &  GPT-2$_{\textup{medium}}$ & GPT-2$_{\textup{large}}$ & GPT-2$_{\textup{xlarge}}$ & GPT-2$_{\textup{xlarge}}$\\
	\hline
    \multirow{5}{*}{\textbf{COPA}} & Baseline & 57.8 & 62.4 & 65.8 & 66.0 & --\\
    & SEQA & 51.4 (63.0) & 53.0 (68.4) & 53.8 (72.0) & 54.4 (75.4)  & 79.4 \\
    & Self-talk  & 59.8 ($\uparrow$2.0)  & 65.0 ($\uparrow$2.6)  & 66.6 ($\uparrow$0.8) & 66.2 ($\uparrow$0.2) & 68.6 \\
    & CGA & -- & -- & -- & -- & 72.2 \\
    & ArT & 60.2 ($\uparrow$2.4) & 64.8 ($\uparrow$2.4) & 67.0 ($\uparrow$1.2) & 67.6 ($\uparrow$1.6) & --\\
    & ArT ($Score_X^i$) & \textbf{61.0} ($\uparrow$3.2) & \textbf{65.6} ($\uparrow$3.2) & \textbf{69.4} ($\uparrow$3.6) & \textbf{69.8} ($\uparrow$3.8) & --\\
    \hline
    \multirow{5}{*}{\textbf{SocialIQA}} & Baseline & 41.3 & 44.3 & 45.5 & 45.9 & --\\
    & SEQA & 34.9 (43.9) & 35.9 (44.6) & 36.5 (46.6) & 36.6 (47.5)  & 47.5 \\
    & Self-talk & 40.5 ($\downarrow$0.8) & 44.8 ($\uparrow$0.5) & 46.1 ($\uparrow$0.6) & 47.2 ($\uparrow$1.3) & 47.5 \\
    & CGA & -- & -- & -- & -- & 45.4 \\
    & ArT & \textbf{42.0} ($\uparrow$0.7) & \textbf{45.6} ($\uparrow$1.3) & \textbf{47.6} ($\uparrow$2.1) & \textbf{47.3} ($\uparrow$1.4) & --\\
    \hline
    \multirow{5}{*}{\textbf{SCT}} & Baseline & 59.6 & 67.4 & 69.1 & 70.5 & --\\
    & SEQA & 50.7 (74.7) & 53.3 (80.5) & 54.2 (82.4) & 54.9 (83.2)  & 83.2 \\
    & Self-talk & 59.8 ($\uparrow$0.2) & \textbf{68.5} ($\uparrow$1.1) & 69.2 ($\uparrow$0.1) & 70.4 ($\downarrow$0.1) & 70.4 \\
    & CGA & -- & -- & -- & -- & 71.5 \\
    & ArT & \textbf{60.2} ($\uparrow$0.6) & 68.3 ($\uparrow$0.9)  & \textbf{69.5} ($\uparrow$0.4) & \textbf{71.6} ($\uparrow$1.1) & --\\
    \bottomrule
	\end{tabular}
	\caption{Accuracy (\%) on COPA, SocialIQA and SCT. All results except last column are run by ourselves. Best results are depicted in boldface (only consider fully unsupervised models for fairness). $\uparrow$/$\downarrow$ refer to relative increase/decrease compared with baseline. For SEQA, we list results of two settings: SRoBERTa$_{\textup{large}}^{Origin}$ (before brackets) and SRoBERTa$_{\textup{large}}^{NLI}$ (in brackets).}
	\label{tab:main_result}
    \end{table*}
    
\subsection{Baseline and Contrastive models}
    Our baseline is constructed as only using PLMs as scorer without any explicit knowledge injection. Formula (\ref{eq:1}) is applied as the scoring function as described in Section \ref{basic}. We also compare ArT with other advanced unsupervised models:
    \begin{itemize}
        \item \textbf{Self-talk}\cite{shwartz2020unsupervised}: It acquires knowledge through a two-stage prompting of PLMs. Different question prefixes had to be specially designed for different tasks.
        \item \textbf{SEQA}\cite{niu2021semantic}: It applies PLMs to generate hundreds of pseudo-answers and compares them with each $option$. However, its scorer relies on PLMs fine-tuned on large-scale labeled NLI datasets, which is not strictly unsupervised. For fair comparison, we design another setting that replaces the fine-tuned PLM with the original one (only pre-trained on unlabeled text) . 
        \item \textbf{CGA}\cite{bosselut2021dynamic}: It employs a generative KB COMET\cite{bosselut2019comet}, which is trained on an existing seed KB (e.g. ConceptNet), to construct context-relevant knowledge graphs to reason over.
    \end{itemize}

\subsection{Setup}
    Following previous work, we employ OpenAI GPT\cite{radford2019language} as the PLM backbone. To reach solid and reproducible results, we conduct experiments on GPT-2 of 4 different scales: distil, medium, large and xlarge. For ArT and Self-talk, the same scale GPT is applied during both knowledge generating and $option$ scoring. For SEQA, GPTs of different scales are used for pseudo-answers generation and SRoBERTa$_{\textup{large}}$\cite{reimers2019sentence} is used for semantic similarity calculation. To distinguish, SRoBERTa$_{\textup{large}}^{NLI}$ and SRoBERTa$_{\textup{large}}^{Origin}$ refer to 
    SRoBERTa$_{\textup{large}}$ with and without further fine-tuning on NLI datasets, respectively. For Self-talk\footnote{ \url{https://github.com/vered1986/self_talk}} and SEQA\footnote{ \url{https://github.com/heyLinsir/Semantic-based-QA}}, we re-run their codes with their original settings and report both our re-running results\footnote{For each setting except GPT-2$_{\textup{xlarge}}$ (limited by the computational power), we run 3 times and report the average number.} and results coming from their publications. For CGA, we report results provided by \citet{niu2021semantic}. For ArT, we modified the open source code of SIFRank\footnote{https://github.com/sunyilgdx/SIFRank} to enable it to extract more kinds of phrases rather than only noun phrase. The size of notes set ($K$) is set as 32 as default. Except for $Score^i_O$, ArT takes another setting $Score^i_X$ on COPA.

\subsection{Results}
    Table \ref{tab:main_result} shows the results on three benchmarks. The results of our re-running is highly consistent with those reported in their publications (last column). Note that published results on COPA seem to have a deviation with our reproduction. It because that they are on test set, while ours on dev set. \citet{shwartz2020unsupervised} reported 66.0\% on COPA in their paper, which is tested on the dev set of a small version which contains 1/5 instances of that other researches used. And ArT reaches 68.0\% on that set under the same setting.
    
    On all datasets, ArT obtains state-of-the-art performance among almost all fully unsupervised models. Besides, it is noticed that on GPT of different scales, ArT with NT stably brings positive improvement over baseline. On causal reasoning task COPA, adding RT will bring further accuracy improvement.
    
    In contrast with ArT, Self-talk fails to maintain effectiveness, whose accuracy is even slightly lower than baseline from time to time, especially on SCT. It indicates that the knowledge generated by Self-talk could be noisy and as a result it misguides model evaluation.
    
    With the help of SRoBERTa$_{\textup{large}}^{NLI}$, SEQA can reach very impressive results on all the datasets, especially SCT (exceed all models than over 10\%). However, when working in a strictly unsupervised mode, this method quickly becomes invalid (close to random selection). 
    
    We also observe a common phenomenon on all our ArT, baseline and contrastive models: when enlarging GPT-2 from large (750M parameters) to xlarge (1500M parameters), we encounter no obvious model performance increasing and even decreasing in some scenarios. It indicates that there could be a limit to only using the method of increasing model parameters to improve the performance of language models as sentence scorer or knowledge generator.
    
\section{Analysis and Discussion}
\subsection{Effect of Different Modules}
\label{sec:ablation}
    In order to determine the source of performance growth, we conduct ablation study on ArT, as shown in Table \ref{tab:ablation}. As expected, injecting knowledge with our NT has positive effect on all three tasks. When removing the \textit{Rethinking} mechanism, all results slightly decrease, which indicates that \textit{Rethinking} can help increase the quality of generated knowledge. Besides, $Score^i_X$ can bring further improvement whether or not NT is employed on COPA.
    
    \begin{table}[t!]
    \fontsize{10pt}{1.1\baselineskip}
    \selectfont
	\centering
	\begin{tabular}{lccc}  
	\toprule
	\textbf{Models} & \textbf{COPA} & \textbf{SocialIQA} & \textbf{SCT} \\
	\hline
    Baseline & 65.8 & 45.5 & 69.1 \\
    \quad$+Score^i_X$ & 68.9 & -- & -- \\
    \hdashline
    \quad$+$NT & 67.0 & \textbf{47.6} & \textbf{69.5} \\
    \quad$+$NT$-\textit{Rethinking}$ & 66.2 & 46.7 & 69.4 \\
    \quad$+$NT$+Score^i_X$ & \textbf{69.4} & -- & -- \\
    \bottomrule
	\end{tabular}
	\caption{Ablation study for ArT modules on GPT-2$_{\textup{large}}$.}
	\label{tab:ablation}
    \end{table}
    
\subsection{Number of Notes and Keyphrases}
    \label{sec:notesize}
\begin{figure}[htb]
	\centering
	\includegraphics[width=0.42\textwidth]{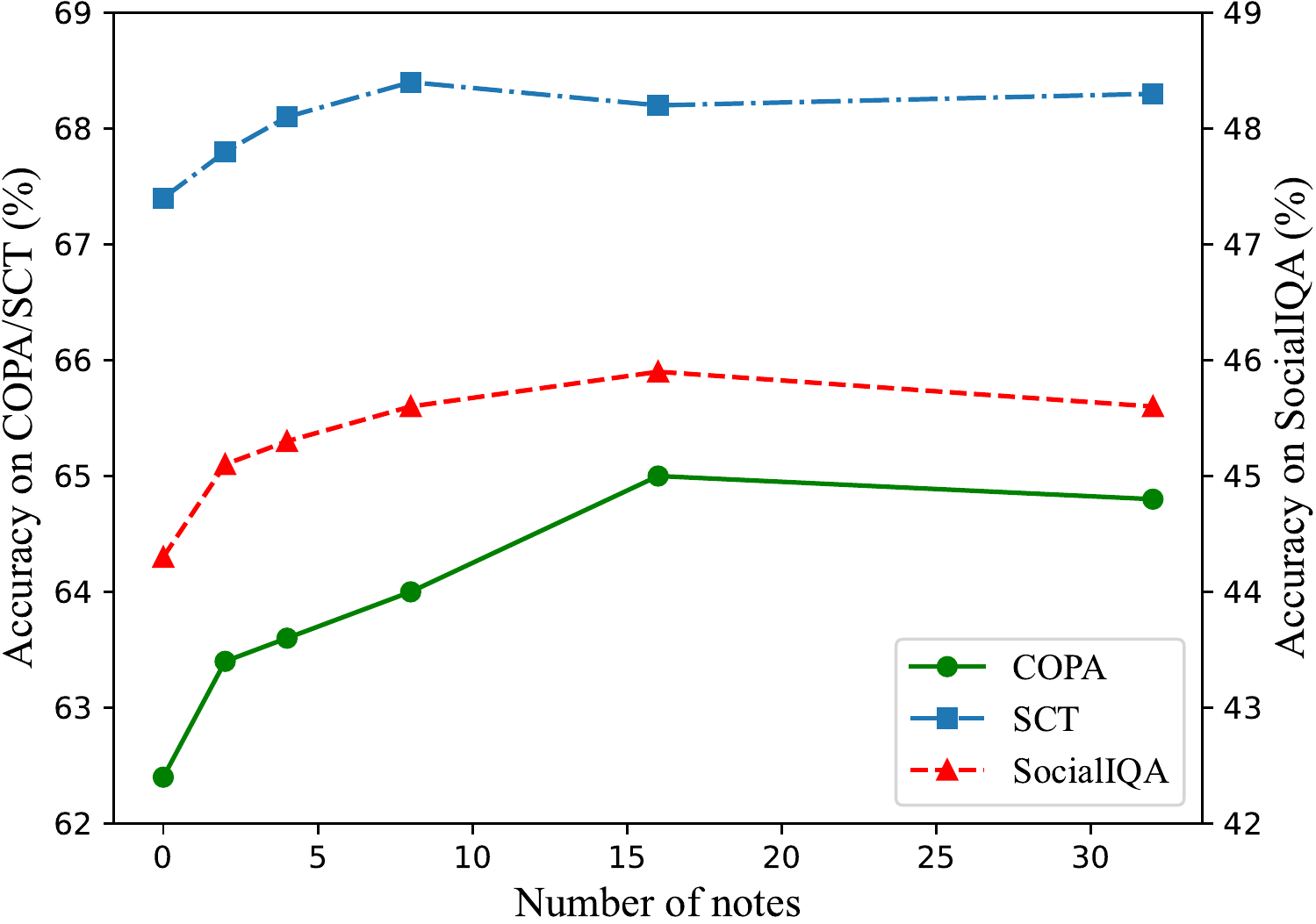}
	\caption{Accuracy curves of ArT on GPT-2$_{\textup{medium}}$ w.r.t the size $K$ of notes set.}
	\label{fig:notes_size}
    \end{figure}

   Figure \ref{fig:notes_size} shows the effect of notes set size $K$. On all the benchmarks, along with the increasing of $K$, the accuracy curves basically show an upward trend, which indicates that as the number of generated knowledge increases, ArT will not accumulate too much noise as \citet{niu2021semantic} observed in Self-talk. Therefore, compared with Self-talk, ArT tends to generate highly related knowledge, which is contributed by our NT mechanism.
   
    \begin{table}[htb]
    \fontsize{10pt}{1.1\baselineskip}
    \selectfont
	\centering
	\begin{tabular}{lccccc}  
	\toprule
	\textbf{$N$} & 1 & 3 & 5 & 7 & 9 \\
	\hline
	baseline & \multicolumn{5}{c}{59.6} \\
	ArT & 59.6  & \textbf{60.2}  & \textbf{60.2}  & 60.0 & 60.0 \\
    \bottomrule
	\end{tabular}
	\caption{Accuracy (\%) of ArT (GPT-2$_\textup{distil}$) on SCT under different settings of $N$.}
	\label{tab:numofkw}
    \end{table}
    In our experiments (Section \ref{sec:experiment}), the number ($N$) of keyphrases to extract is set as 5 as default. To show the effect of $N$, we conduct an ablation study on SCT by setting $N \in \{1, 3, 5, 7, 9\}$. The PLM is selected as GPT-2$_\textup{distil}$. Table \ref{tab:numofkw} shows the results. It is noticed that extracting more keyphrases does not always results in a better performance. But in general, the choice of $N$ does not have an obvious impact on the final performance. 
    
    \begin{table}[h!]
    \fontsize{10pt}{1.1\baselineskip}
    \selectfont
	\centering
	\begin{tabular}{llcc}  
	\toprule
	\textbf{Dataset} & \textbf{Model} & \textbf{Rationality} & \textbf{Usefulness} \\
	\hline
    \multirow{2}{*}{COPA} & Self-talk & 0.24 & 0.20 \\
    & ArT & \textbf{0.32} & \textbf{0.28} \\
    \hdashline
    \multirow{2}{*}{SocialIQA} & Self-talk & 0.17 & 0.16 \\
    & ArT & \textbf{0.26} & \textbf{0.27} \\
    \bottomrule
	\end{tabular}
	\caption{Human evaluation on the rationality and usefulness of generated knowledge.}
	\label{tab:human_eval}
    \end{table}
    
    \subsection{Quality of Generated Knowledge}
    To compare the quality of generated knowledge (whether the knowledge is reasonable enough to be a ``fact'' and correlative enough to be useful), we conduct human evaluation on the rationality (-1: \textit{irrational}, 0: \textit{meaningless}, 1: \textit{rational}) and usefulness (-1: \textit{negative}, 0: \textit{neutral}, 1:
    \textit{positive}) of knowledge generated by ArT and Self-talk.
    Two annotators are asked to independently annotate 100 randomly selected knowledge for both COPA and SocialIQA. The overall average scores of two annotators for each dataset are shown in Table \ref{tab:human_eval}. It is noticed that ArT outperforms Self-talk in generating knowledge with both rationality and usefulness.

    \begin{figure}[t!]
	\centering
	\includegraphics[width=0.45\textwidth]{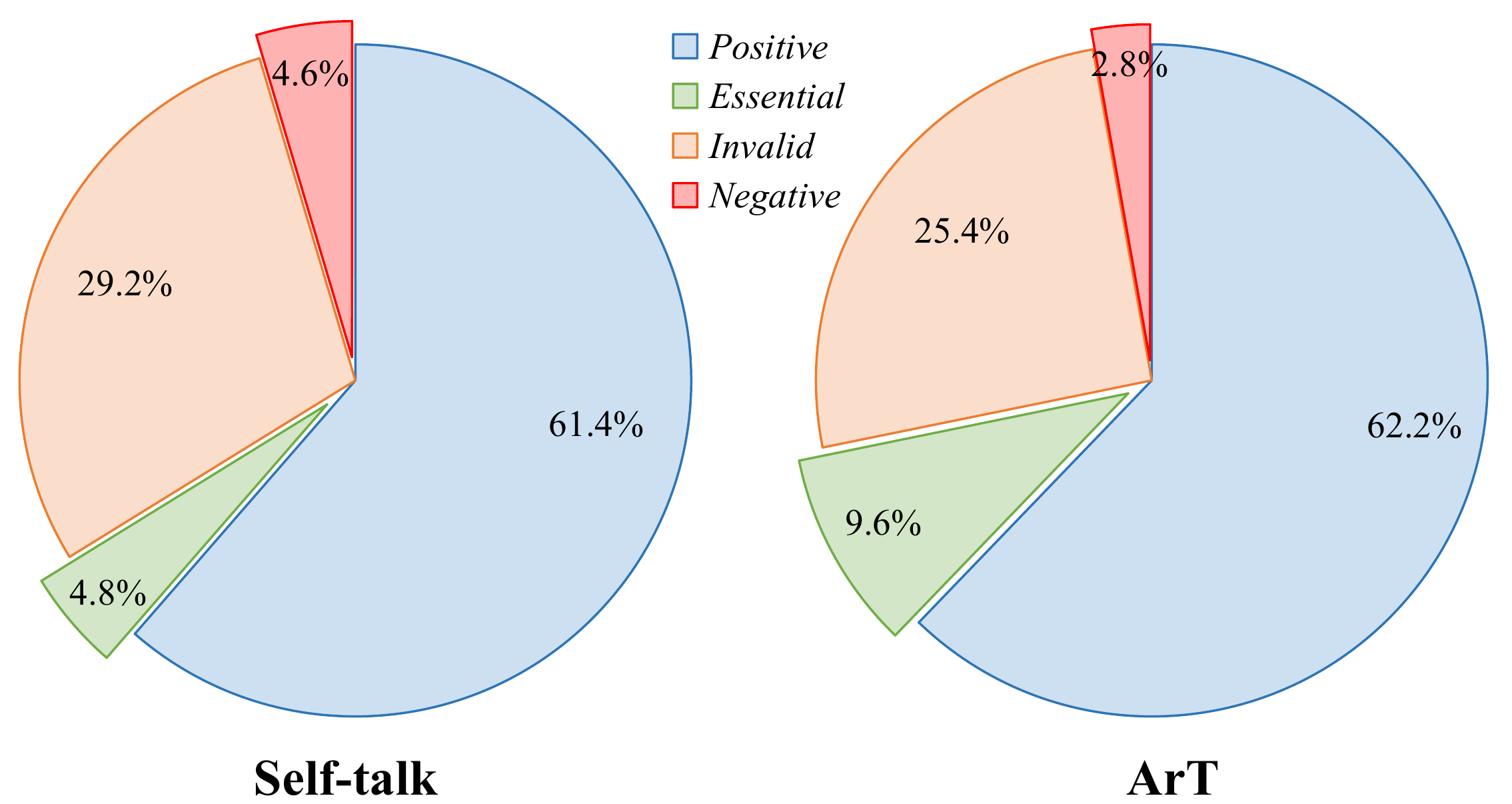}
	\caption{Statistics on the quality of knowledge generated by Self-talk and ArT on COPA.}
	\label{fig:pie}
    \end{figure}
    
    We further make statistics on the kinds of all the knowledge generated by Self-talk and ArT on COPA. We divide them into four classifications:
    
    \begin{itemize}
        \item \textit{Positive}: Baseline makes the right prediction. Adding the knowledge still makes a right prediction.
        \item \textit{Essential}: Baseline makes the wrong prediction. Adding the knowledge helps make a right prediction. 
        \item \textit{Invalid}: Baseline makes the wrong prediction. Adding the knowledge still makes a wrong prediction.
        \item \textit{Negative}: Baseline makes the right prediction. Adding the knowledge leads to a wrong prediction.
    \end{itemize}
    
    As shown in Figure \ref{fig:pie}, by doubling \textit{Essential} and reducing \textit{Negative} knowledge, ArT outperforms Self-talk in generating high-quality commonsense.
    
    Figure \ref{fig:knowledge_example_1} and \ref{fig:knowledge_example_2} in Appendix show some examples of knowledge generated by ArT, Self-talk and SEQA for COPA, SocialIQA and SCT. As can be seen, ArT generates highly related knowledge on the basis of focusing on keyphrases of given context. While Self-talk may generate meaningless or even noisy knowledge. It is also noticed that SEQA could generate very reasonable pseudo-answers. And to distinguish the relationship between these pseudo-answers and given $options$ relies much on a sentence embedding extractor fine-tuned on labeled NLI data.
    
    \subsection{Effect of Note Types}
    As we designed three types of notes: NP, VP and PNP, we only consider one type at one time to show the effect of each in different benchmarks, as shown in Table \ref{tab:note_types} (The PLM is GPT-2$_\textup{medium}$). 
    
    \begin{table}[h!]
    \fontsize{10pt}{1.1\baselineskip}
    \selectfont
	\centering
	\begin{tabular}{llccc}  
	\toprule
	\textbf{Dataset} & \textbf{NP} & \textbf{VP} & \textbf{PNP} & \textbf{All / None}\\
	\hline
    COPA & 63.6 & 63.2 & 62.4 & 64.8 / 62.4 \\
    SocialIQA & 45.0 & 44.8 & 45.3 & 45.6 / 44.3 \\
    SCT & 68.0 & 67.9 & 68.1 & 68.3 / 67.4 \\
    \bottomrule
	\end{tabular}
	\caption{Effect of each type of notes on different tasks.}
	\label{tab:note_types}
    \end{table}
    
    It is noticed that each type of notes has a positive effect when applied separately, and their combination works better. Note that COPA has no person name phase (PNP), so PNP notes does not work on it. On SocialIQA, PNP works best among three types of notes. This is reasonable since SocialIQA focus much on human behavior in social interactions.
    
    \section{RT for Other Questions}
    \label{sec:rt_ablation}
    Although RT is designed to enhance causal reasoning, we also explored if RT has the potential to help other questions. To investigate this, we apply it on other two datasets: SocialIQA and SCT. Note that in these tasks the opposite question is hard to define, therefore we simply exchange the position of $option$ and $context$, that is $ST^i_O=<C, Q, O^i>$ and $ST^i_R=<O^i, Q, C>$. We employ three scoring functions on basis of GPT-2$_\textup{medium}$ baseline, as shown in Table \ref{tab:rt}. 
    
    \begin{table}[htb]
    \fontsize{10pt}{1.1\baselineskip}
    \selectfont
    \centering
    \begin{tabular}{lccc}  
    \toprule
    \textbf{Functions} & \textbf{COPA} & \textbf{SocialIQA} & \textbf{SCT} \\
    \hline
    $Score^i_O$ & 62.4 & \textbf{44.3} & \textbf{67.4} \\
    $Score^i_R$ & 63.2 & 42.5 & 62.8 \\
    $Score^i_X$ & \textbf{65.3} & 44.1 & 65.4 \\
    \bottomrule
    \end{tabular}
    \caption{Accuracy (\%) of GPT-2$_{\textup{medium}}$ baseline with different scoring functions.}
    \label{tab:rt}
    \end{table}
    
    On all the tasks, $Score^i_O$ and $Score^i_R$ can obtain positive results (much better than random selection). By simply averaging ($Score^i_X$), on COPA it can reach a higher score. On other tasks, it falls into the middle of $Score^i_O$ and $Score^i_R$. Considering that $Score^i_R$ shows a comparable performance with $Score^i_O$ by simply exchanging the position of $option$ and $context$, developing a general method for opposite question definition or designing a more exquisite method to integrate $Score^i_O$ and $Score^i_R$ perhaps could make RT suitable for questions beyond causal reasoning, which will be the key point of our future work.
    
\section{Conclusion}
    Commonsense QA has been a challenging task for it requires extra knowledge beyond the given context. In consideration of the high resource consumption of building knowledge bases (KBs) and the rarity of high-quality labeled data, this work aims at addressing commonsense QA in a fully KBs-free and unsupervised way. Inspired by the association process of human thinking, we propose \textbf{A}ll-\textbf{r}ound \textbf{T}hinker (ArT), which first focuses on key parts in the given context, and then generates highly related knowledge on such a basis in an association way. Besides, a reverse thinking mechanism is introduced to further enhance bidirectional inferring for causal reasoning as human will do. We test ArT on three benchmarks: COPA, SocialIQA and SCT. ArT outperforms previous advanced unsupervised models and shows stable performance on all scales of PLM backbones.
\bibliography{ArT}
\bibliographystyle{acl_natbib}

\clearpage
\appendix
\begin{figure*}[h!]
	\centering
	\includegraphics[width=0.99\textwidth]{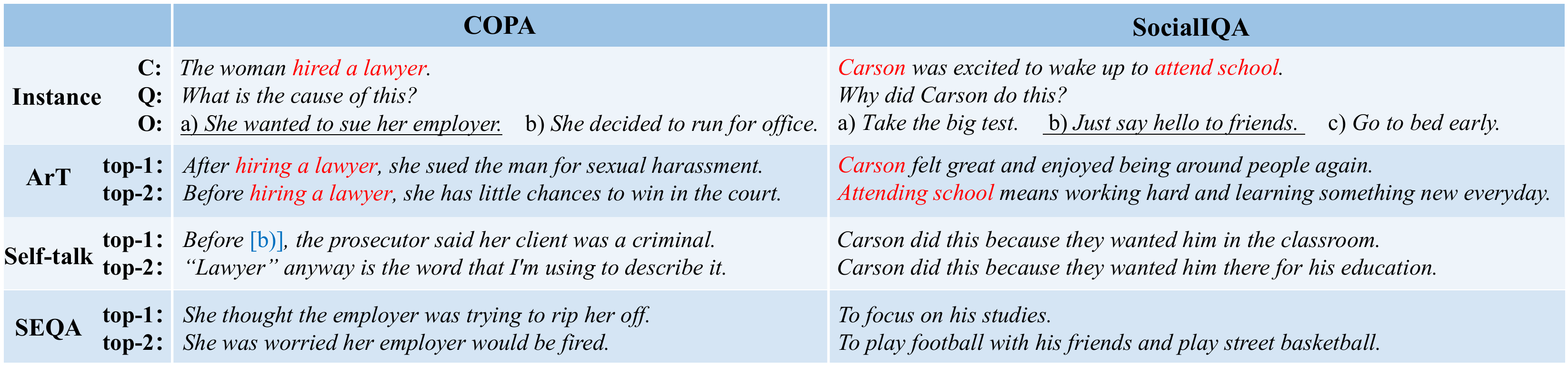}
	\caption{Top two most contributing generated knowledge for instances of COPA and ScocialIQA. The correct options are underlined. Keyphrases extracted by ArT are marked in red.}
	\label{fig:knowledge_example_1}
\end{figure*}
\begin{figure*}[h!]
	\centering
	\includegraphics[width=0.75\textwidth]{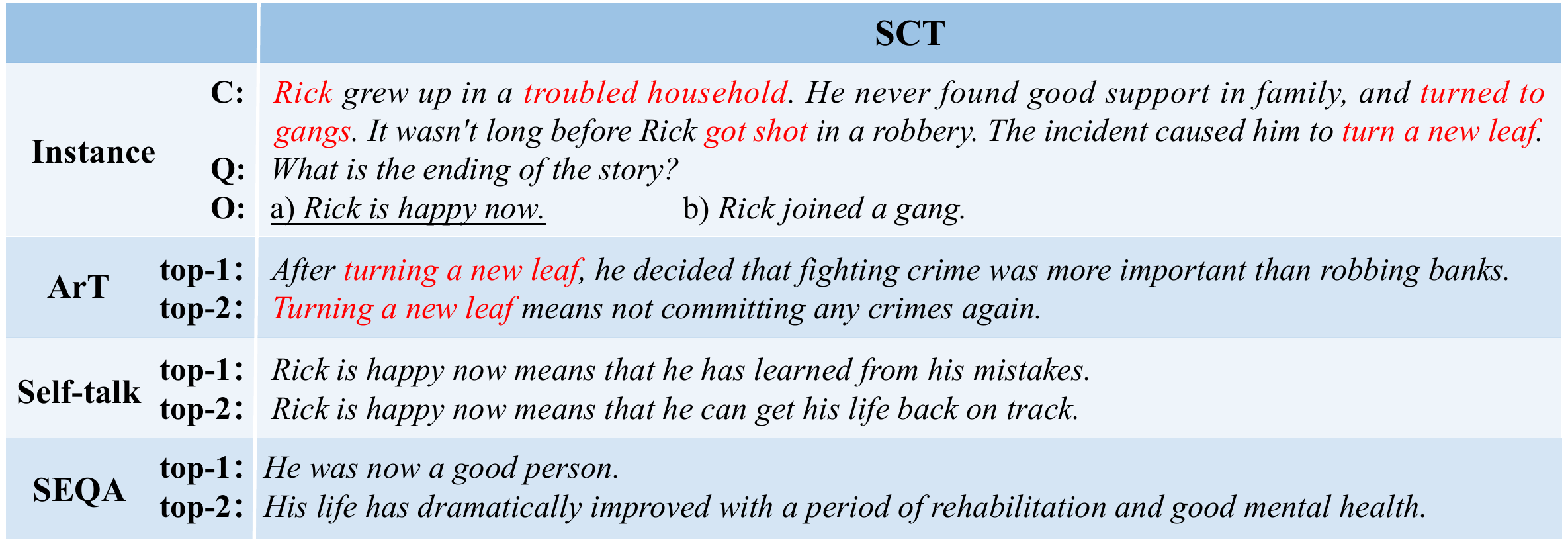}
	\caption{Top two most contributing generated knowledge for instances of SCT. The correct options are underlined. Keyphrases extracted by ArT are marked in red.}
	\label{fig:knowledge_example_2}
\end{figure*}
\end{document}